\documentclass[11pt,a4paper]{article}
\usepackage[hyperref]{emnlp2020}
\usepackage{times}
\usepackage{latexsym}

\usepackage{microtype}

\usepackage{url}
\usepackage{enumitem}
\usepackage{subfig}
\usepackage{booktabs}
\usepackage{makecell}
\usepackage{graphicx}

\newcommand*\samethanks[1][\value{footnote}]{\footnotemark[#1]}

\newcommand{\minisection}[1]{\noindent{\bf #1}\hspace{0.6em}}

\aclfinalcopy 
\def\aclpaperid{607} 

\title{Reasoning about Goals, Steps, and Temporal Ordering with WikiHow}
\author{Li Zhang\thanks{\hspace{4pt} Equal contribution.} \\\And
  Qing Lyu\samethanks \\
  University of Pennsylvania \\
  {\tt \{lyuqing,zharry,ccb\}@seas.upenn.edu} \\\And
  Chris Callison-Burch \\}
\date{}

\begin{document}
\maketitle

\begin{abstract}
 We propose a suite of reasoning tasks on two types of relations between procedural events: \textsc{goal-step} relations (``learn poses'' is a step in the larger goal of ``doing yoga'') and \textsc{step-step temporal} relations (``buy a yoga mat'' typically precedes ``learn poses''). We introduce a dataset targeting these two relations based on wikiHow, a website of instructional how-to articles. Our human-validated test set serves as a reliable benchmark for commonsense inference, with a gap of about 10\% to 20\% between the performance of state-of-the-art transformer models and human performance. Our automatically-generated training set allows models to effectively transfer to out-of-domain tasks requiring knowledge of procedural events, with greatly improved performances on SWAG, Snips, and Story Cloze Test in zero- and few-shot settings.
\end{abstract}

\section{Introduction}
If you ask Alexa or Siri where to ``buy a yoga mat,'' it should ideally infer that your goal is probably to ``do yoga'' and therefore suggest information on subsequent steps like ``learn some poses.'' This requires the system to reason about the \textsc{goal-step} relation and the \textsc{step-step temporal} relation among events in a procedure. Though event relation reasoning is a popular task, existing datasets mostly focus on temporal relations \cite{pustejovsky2003timebank,  ning-etal-2018-multi}, causal relations \cite{hashimoto2014toward, caselli2017event}, spatiotemporal containment and coreference relations \cite{glavavs2014hieve, liu2014supervised}. Less attention has been paid to relations among procedural events, the understanding of which is critical to task-oriented intelligent systems. The knowledge of procedural events is also crucial to learning scripts \cite{feigenbaum1981handbook}, which describe sequences of stereotypical human activities. 

\begin{figure}[t!]
    \centering
    \includegraphics[scale=0.58]{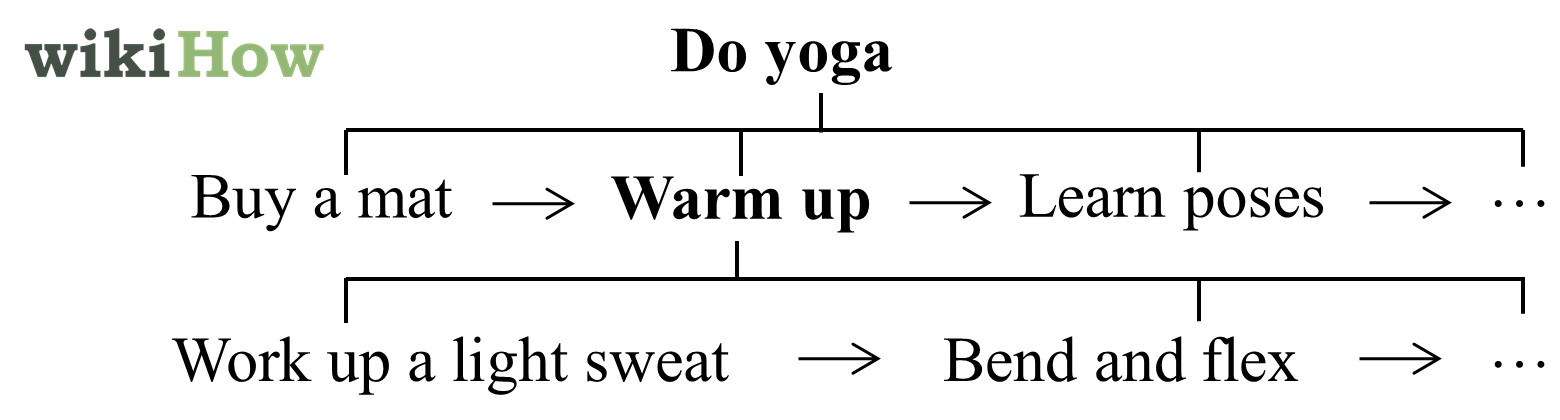}
    \caption{Goals and steps (slightly paraphrased) from wikiHow articles ``How to Do Yoga'' and ``How to Warm Up''. The lines denote \textsc{goal-step} relations; the arrows denote \textsc{step-step temporal} relations.}
    \label{figure:example}\vspace{-.45cm}
\end{figure}

To bridge this gap, we propose a dataset for \textit{goal-step inference} targeting these two event relations. We collect data from wikiHow\footnote{\url{wikihow.com}}, a website consisting of more than 110,000 professionally-edited how-to articles spanning a surprisingly wide range of domains. Each wikiHow article describes a commonplace human activity, organized as a goal and a sequence of steps (Figure~\ref{figure:example}). Our dataset includes 3 tasks: inferring steps given a goal, the goal given a step, and the ordering between two steps given a goal. For each task, we automatically generate 100,000 to 800,000 examples as the training set, using a negative sampling strategy based on semantic similarity; we also provide a human-validated test set with 1,000 to 3,000 examples.\footnote{The data and models are available at \url{https://github.com/zharry29/wikihow-goal-step}.}

Our test set serves as a reliable benchmark for commonsense inference, with a performance gap of 10\% to 20\% between human and state-of-the-art transformer models trained in-domain. Moreover, when pre-trained on our tasks, a model can transfer knowledge of procedural event relations to other NLU tasks, with a zero-shot improvement over the baselines by 24\% for a commonsense reasoning benchmark \cite{zellers2018swagaf}, 13\% for a story cloze test \cite{mostafazadeh-etal-2016-corpus} and 64\% for an intent detection task \cite{coucke2018snips}.

\section{WikiHow Corpus}

We construct a new corpus by crawling the latest wikiHow website. Our corpus has 112,505 how-to articles after deduplication in an easy-to-process JSON format (statistics by category are shown in Appendix~\ref{appendix:category_distribution}). Each article contains: main bodies of texts (titles, methods/parts, headers, descriptions), related articles, references, Q\&A, tips and warnings. To facilitate multi-modal research, we also include links to images and videos aligned with texts. 

\section{Goal-Step Inference Tasks}
We propose 3 goal-step inference tasks derived from the corpus. In each article, we define \textbf{Goal} as the title without ``How to'', and \textbf{Step} as the header of each paragraph (example shown in Figure~\ref{figure:example}). 

\subsection{Step Inference Task}\label{step-inference-task}
We first introduce the \textit{Step Inference} task, targeting \textsc{goal-step} relations between events. We formulate this as a 4-choose-1 multiple choice format evaluated using accuracy.

In this task, a system is given a prompt goal and 4 candidate steps and needs to choose the step that helps achieve the goal. For example, given the goal ``Prevent Coronavirus'' and the candidate steps:\\
A. wash your hands \hspace{20pt} B. wash your cat \\
C. clap your hands \hspace{24pt} D. eat your protein\\
the correct step would be A. 

Obtaining the prompt and the positive candidate is straightforward, as we sample them iteratively from each how-to article. However, it is challenging to sample negative candidates \cite{chao-etal-2018-negative,zellers2019vcr} which should have high semantic relatedness with the positive candidate (or the question becomes trivial) while being incorrect answers. We first map each step in wikiHow to a vector representation by taking the average of the BERT embeddings \cite{devlin-etal-2019-bert} of the verbs. Given the positive step, we then choose 3 steps under different wikiHow categories with the highest cosine similarity to it as the negative candidates (see Appendix~\ref{appendix:negative} for other strategies). The nearest-neighbors are computed using FAISS \cite{JDH17}.

It has recently become clear that the latest NLP models can exploit statistical artifacts from a dataset \cite{poliak-etal-2018-hypothesis,Si2019WhatDB,zellers-etal-2019-hellaswag}. To prevent the model from learning the negative sampling strategy and relying on just the candidates, we randomly reassign one of the candidates as positive, and the others as negative. Then, we replace the prompt goal with the goal attached to the new positive candidate. This strategy ensures that any model performs no better than chance when given access to only the candidates and not the prompt. 

For each step in wikiHow, we create an example by using it as the positive candidate, followed by the negative sampling and label reassignment processes as described above. Then, we apply a collection of hand-crafted filters to remove low-quality examples (Appendix~\ref{appendix:filters}).

\subsection{Goal Inference Task}
\label{goal-inference-task}
Next, we introduce the \textit{Goal Inference} task, formulated in a similar way as Step Inference. 

In this task, a system is given a prompt step and 4 candidate goals and needs to choose the correct goal which the step helps achieve. For example, given the step ``choose a color of lipstick'' and the candidate goals:\\
A. Get Pink Lips \hspace{18pt} B. Read One's Lips \\
C. Lip Sync \hspace{39pt} D. Draw Lips\\
the correct goal would be A. 

For each goal in wikiHow, we create the set of 4 candidates by using it as the positive candidate, followed by the negative sampling, label reassignment, and filtering processes as in Step Inference. For each positive candidate goal, we use each of its steps to create an example. 

\subsection{Step Ordering Task}
Finally, we introduce the \textit{Step Ordering} task, targeting \textsc{step-step temporal} relations between events. This task is in a 2-choose-1 multiple choice format evaluated using accuracy.

In this task, given a prompt goal and 2 steps, a system needs to determine which step temporally precedes the other. For example, given the goal ``Clean Silver'' and the steps:\\
A. dry the silver \hspace{10pt} B. handwash the silver\\
the correct answer would be B precedes A.

Unfortunately, not all steps in every wikiHow article follow an underlying order. We observe that there are 2 types of wikiHow articles. One is \textit{unordered}, where the steps are parallel alternatives, such as ways to ``Stay Healthy'' (``develop an exercise routine", ``get enough sleep", ``eat a healthy diet'', etc.). The other is \textit{ordered}, such as recipes for cooking or manuals for fixing appliances, where most steps should be taken sequentially. 

We ask 3 annotators to label 1,000 wikiHow articles as ordered or not as a coarse-grained approximation for whether their steps are ordered. We finetune a pre-trained RoBERTa model using 5-fold cross-validation, finding an average precision of 88\%. We then ask a 4\textsuperscript{th} annotator to label another 40 articles as the held-out test set, where the finetuned model achieves 100\% precision. Finally, we only consider articles that the model predicts as ordered (around 40\%) for the Step Ordering task. 

For each goal in wikiHow, we create a set of examples by using it as the prompt and sampling every pair of its adjacent steps as candidates. Then, we randomly shuffle the candidates, so each appears first with 50\% chance. 

\subsection{Test Set Construction and Validation}\label{crowdsourcing-validation}
There exists some noise in our automatically generated examples, because some of them do not have a single correct answer. Errors can be introduced when a sampled negative candidate is in fact correct. For example, in the Goal Inference task, consider an example where the give step is ``practice swings'', the expected positive candidate step is ``Play Golf'', and a candidate negative example is ``Play Drums''. ``Play Drums'' is sampled due to its high embedding similarity with ``Play Golf'' and is also a reasonable goal for ``practice swings (of the drumsticks)''. This is an ambiguous example and should be excluded from the test set. Therefore, we ask crowd workers to validate a subset of the examples. 

We perform crowdsourcing on Amazon Mechanical Turk, requiring Master Qualification and a lifetime HIT approval rate over 90\%.\footnote{HIT designs and related details are in Appendix~\ref{appendix:crowdsourcing-details},~\ref{appendix:subbenchmarks}.} 

For each of Step Inference and Goal Inference, we randomly sample 4,800 examples as input, and for each example we ask 3 crowd workers to choose the most likely candidate. Every HIT includes 15 examples with a pay of \$0.83, estimated to be completed in 5 minutes, equivalent to an hourly rate of \$9.96.

For Step Ordering, we randomly sample 9,300 examples, and for each example we ask 3 crowd workers to order the events (with a ``neutral'' option). Every HIT includes 30 examples with a pay of \$0.83, estimated to be completed in 5 minutes, equivalent to an hourly rate of \$9.96. 

In the test set, we only keep examples where all 3 workers agree with the gold label as our benchmark. We remove all examples from the automatically generated ones whose prompt or candidates appear in the test set, and use the remaining data as the training set.

\section{In-Domain Evaluation}
\label{model_perf}
We finetune pretrained BERT \cite{devlin-etal-2019-bert}, XLNet \cite{yang2019xlnet}, RoBERTa \cite{DBLP:journals/corr/abs-1907-11692} and GPT-2 \cite{radford2019language} models on the training set and report accuracy on the test set. Modeling details including hyperparameter settings are shown in Appendix~\ref{appendix:modeling_details}. To benchmark human performance, two authors each annotate 100 random examples from the test set and report the average accuracy. The results are shown in Table~\ref{performance}, indicating a performance gap of 10\% to 20\% between human and models trained on all available in-domain data. 

\begin{table}[t!]
\centering
\begin{tabular}[t]{lllll}
\toprule
 & \makecell{Step\\Infer.} & \makecell{Goal\\Infer.} & \makecell{Step\\Ordering} \\ \midrule
Train size & 374,278 & 185,231 & 836,128 \\
Test size & 2,250 & 1,703 & 3,100 \\ \midrule
BERT & .874 & .798 & .819 \\
XLNet & .867 & .783 & .826 \\ 
RoBERTa & .882 & .820 & .835 \\
GPT-2 & .836 & .686 & .801 \\ 
\midrule
Human & .965 & .980 & .975 \\
\bottomrule
\end{tabular}
\caption{The accuracy of state-of-the-art models on the test sets after being finetuned on our training sets.}
\label{performance}
\end{table}

\subsection{Open-Ended Examples}\label{open-ended}
In addition to quantitatively evaluating models on our multiple-choice tasks, we perform qualitative evaluation on some open-ended examples from wikiHow unseen during training, using RoBERTa.\footnote{Modeling details and more examples are in Appendix~\ref{appendix:qualitative}.}

For Step Inference, we rank 100 steps with high embedding similarity for their likelihood of helping achieve a given goal. For example, for the goal ``Eat in Islam'', the top 3 ranked steps are ``understand what type of meats are permissible'' (correct), ``start by adding mild spices to your food,'' and ``gather supplies and ingredients.'' Similarly for Goal Inference, we rank 100 goals against some steps. For example, for the steps ``spend the holiday with your beloved, eat KFC, check out the light displays,'' the top 3 ranked goals are ``Celebrate a Japanese Christmas'' (correct)\footnote{KFC and light displays are Japanese Christmas traditions \cite{kimura2005christmas}.}, ``Celebrate a Czech Christmas,'' and ``Celebrate a British Christmas.'' These examples show that the model trained on our data can retrieve texts based on \textsc{goal-step} relations, beyond simply semantic relevance. 

For Step Ordering, the model can perfectly order some articles with as much as 10 steps. For example, given the goal ``Clean a Trumpet,'' the first 5 predicted, ordered steps are ``gather your materials,'' ``disassemble your trumpet,'' ``fill up a bathtub,'' ``place a towel down in the tub,'' and ``set your trumpet parts down to soak.'' This shows that the model trained on our data can order certain long sequences of events based on \textsc{step-step temporal} relations.

\section{Out-of-Domain Transfer Learning}

\begin{figure}[t!]
    \centering
    \includegraphics[scale=0.4]{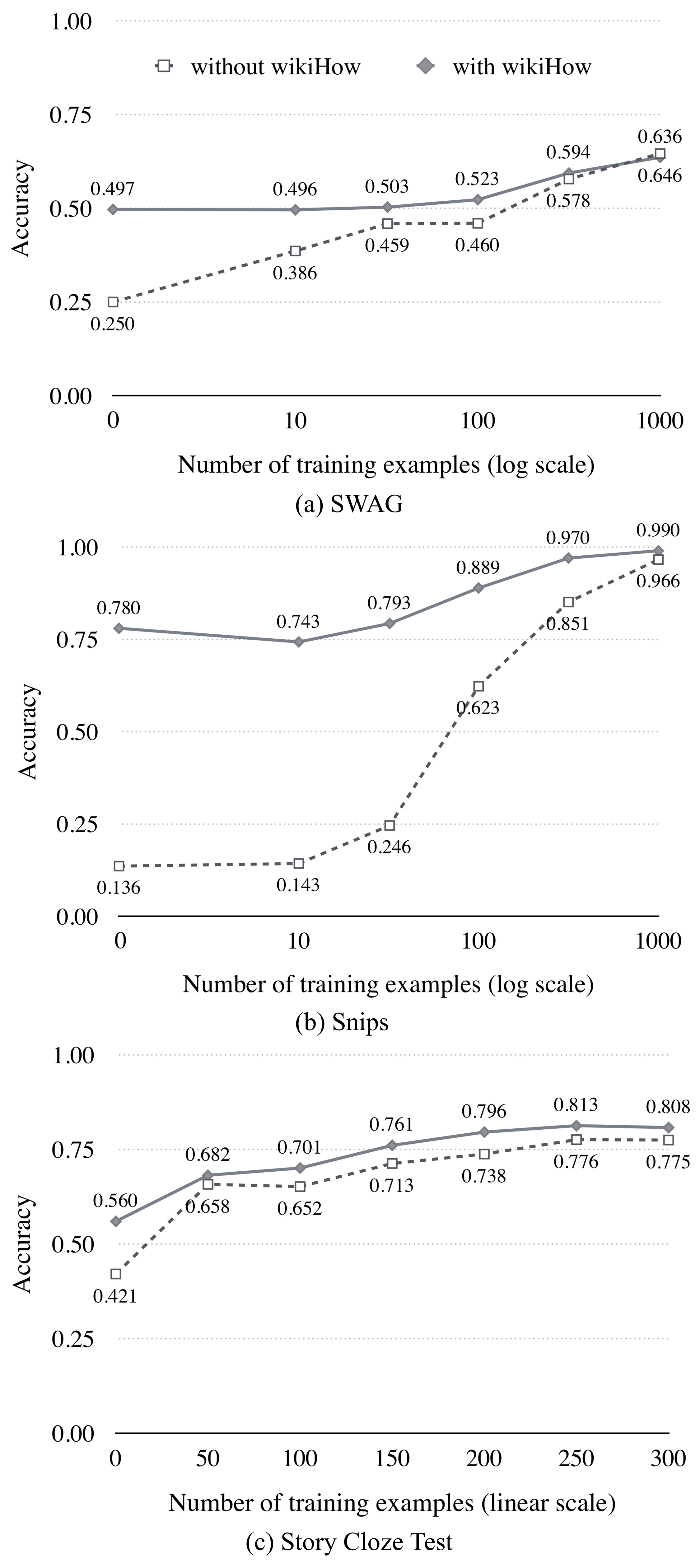}
    \caption{Accuracy of RoBERTa on SWAG, Snips and Story Cloze Test with different training set sizes, with and without being previously fine-tuned on our tasks. \vspace{-0.45cm}}
    \label{transfer-curve}
\end{figure}

To show that our tasks can serve as an effective transfer learning resource especially in zero- or few-shot settings, we consider 3 tasks in different domains, using a subset of their training data to simulate a low-resource scenario. Therefore, we are not comparing to the state-of-the-art performances involving the entire in-domain training sets.

For each target task, we finetune a vanilla RoBERTa model and one pretrained on our task on increasingly larger portions of the target training set, and observe accuracy on the validation set, as the test set labels are not publicly available.\\
\textbf{SWAG} \cite{zellers2018swagaf} is a \textit{commonsense inference} dataset in the video caption domain. Given a context, a system chooses one event most likely to happen from four candidates. For transfer learning, we use up to 1,000 examples for training and the standard validation set. We use the model trained on our Step Inference task to transfer to this task. \\
\textbf{Snips} \cite{coucke2018snips} is an \textit{intent detection} dataset in the dialogue system and spoken query domain, where a system classifies an utterance into one of 7 intents. For transfer learning, we use up to 1,000 examples for training and the standard validation set. We use the model trained on our Goal Inference task to transfer to this task. To enable zero-shot transfer, we convert each example in our training data to a 7-choose-1 format by adding 3 empty strings as additional negative candidates. \\
\textbf{Story Cloze Test} \cite{mostafazadeh-etal-2016-corpus} is a \textit{story understanding} dataset in the fiction domain, where a system chooses an ending to a 4-sentence-story from 2 candidates. We use up to 314 examples for training and 1,571 examples for validation, from the 2016 and 2018 data releases after removing duplicates. We use the model trained on our Step Ordering task to transfer to this task. To mimic the ``next sentence prediction'' format, we convert each example in our task to a ``next step prediction'' question with 4 prompt steps and 2 candidate steps, exactly one of which happens after the prompt.

Figure~\ref{transfer-curve} shows the learning curves of the downstream tasks with an increasing number of their training samples, demonstrating a clear advantage of using our training data in low-resource settings. For SWAG, the model trained on our data has a zero-shot performance 24\% over chance, outperforming the vanilla model when up to 1,000 training examples are given. For Snips, the model trained on our data boasts an impressive 78\% zero-shot performance, approaching perfect accuracy rapidly after some in-domain training. For the Story Cloze Test which has the largest domain-mismatch with our data, the model still benefits from the knowledge learned from it consistently, given any portion of in-domain training data up to the full size in our experiment. These results show that the model learns real-world procedural knowledge from our wikiHow-based tasks, which can be readily applied to various domains and writing styles. 

\section{Related Work}
\minisection{Script Learning} A field of research related to our work is script learning, proposed by \citet{feigenbaum1981handbook}. Scripts encode the knowledge of stereotypical event sequences, such as going to a restaurant or visiting a doctor. A branch of research has focused on distilling \textit{narrative} scripts from newswire and literature \cite{chambers2008unsupervised, jans2012skip, pichotta2014statistical}, while another, which is more similar to our work, focuses on \textit{procedural} scripts that are core to task-oriented intelligent systems. A few large-scale crowdsourced corpora of the latter kind are OMICS \cite{gupta2004common}, SMILE \cite{regneri2010learning}, the \citet{Li2012CrowdsourcingNI} corpus and DeScript \cite{wanzare2016crowdsourced}. As wikiHow articles consist of chains of human activities, we believe wikiHow may be a useful resource for script learning as well. Specifically, while most previous research either mined noisy scripts from raw texts or crowdsourced them, wikiHow’s particular text structure can provide a huge number of clean scripts for free. We will explore it in our future work.

\minisection{WikiHow as a Resource} 
WikiHow has been used in several past NLP efforts, including knowledge-base construction \cite{pareti2014integrating, chu2017distilling}, text generation \cite{nguyen-etal-2017-sequence}, household activity prediction \cite{zhou-etal-2019-learning}, and summarization \cite{Koupaee2018WikiHowAL}. HellaSwag \cite{zellers-etal-2019-hellaswag}, a recent commonsense reasoning dataset, presents a sentence completion task derived from wikiHow texts. However, it is likely that artifacts exist in the dataset, since BERT achieves 41\% accuracy in the candidate-only setting and RoBERTa achieves 83\% zero-shot performance.\footnote{\url{rowanzellers.com/hellaswag/\#leaderboard}} Apart from HellaSwag, \citet{park2018learning} addressed classification tasks involving similar event relations to the ones we consider. Nevertheless, few existing research efforts attempted to prove the potential of wikiHow as a transfer resource on out-of-domain tasks. In comparison, our contributions are two-fold, in that we propose both a human-validated benchmark and an effective learning resource using wikiHow.

\section{Summary}

We propose 3 goal-step inference tasks using wikiHow to complement research of event relation reasoning. Our test sets serve as a reliable benchmark for commonsense inference, and more importantly, our dataset is an effective transfer learning resource, improving transformer models' performance on various tasks in zero- or few-shot settings. This implies a strong potential for pre-training models to better generalize in low-data scenarios.

\section*{Acknowledgments}

This research is based upon work supported in part by the DARPA KAIROS Program (contract FA8750-19-2-1004), and the IARPA BETTER Program (contract 2019-19051600004). The views and conclusions contained herein are those of the authors and should not be interpreted as necessarily representing the official policies, either expressed or implied, of DARPA, IARPA, or the U.S. Government. The U.S. Government is authorized to reproduce and distribute reprints for governmental purposes notwithstanding any copyright annotation therein.

We thank Aditya Kashyap and Arun Kirubarajan for the annotations for article order prediction. We also thank Daphne Ippolito, Reno Kriz, Vivienne B. Chi, Mohammad S. Rasooli, Rebecca Iglesias-Flores, Eleni Miltsakaki, Liam Dugan, Jian Hu, Muhao Chen, Elior Sulem, Hongming Zhang, Hangfeng He, Sha Li, Heng Ji, Dan Roth
and the anonymous reviewers for their valuable feedback. 

\bibliographystyle{acl_natbib}
\bibliography{emnlp2020}

\appendix

\section{Modeling Details}\label{appendix:modeling_details}

All our models are implemented using the HuggingFace Transformer service\footnote{\url{https://github.com/huggingface/transformers}}. 

We tune our model hyperparameters using cross-validation on our benchmarks. We do so on the validation sets of the out-of-domain datasets. As 4 different models each with different hyperparameters are involved, we do not list them here. Instead, the hyperparameter values and pretrained models are available in our Github repository. We save the model every 1,000 training steps, and choose the model with the highest validation performance to be evaluated on the test set. 

We run our experiments on an NVIDIA GeForce RTX 2080 Ti GPU, with half-precision floating point format (FP16) with O1 optimization.

\section{Category Distribution of WikiHow Articles}\label{appendix:category_distribution} 
\begin{figure}[t!]
    \centering
    \includegraphics[scale=0.28]{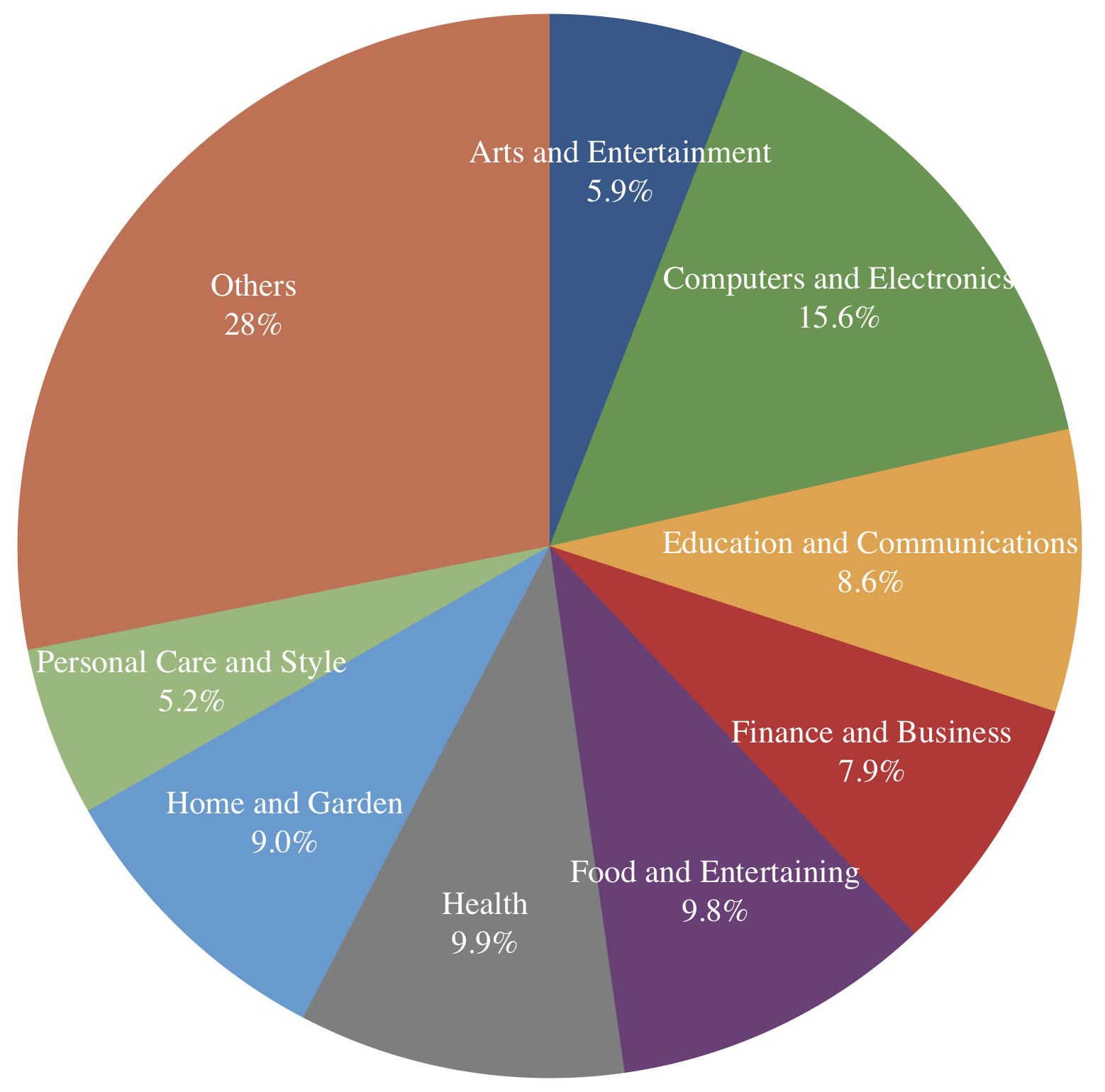}
    \caption{Category distribution of wikiHow articles.}
    \label{figure:cat_dist_app}
\end{figure}

WikiHow has articles from a broad range of domains, with 19 top-level categories: Arts and Entertainment, Cars \& Other Vehicles, Computers and Electronics, Education and Communications, Family Life, Finance and Business, Food and Entertaining, Health, Hobbies and Crafts, Holidays and Traditions, Home and Garden, Personal Care and Style, Pets and Animals, Philosophy and Religion, Relationships, Sports and Fitness, Travel, Work World, and Youth. We plot the distribution of the top eight categories in Figure~\ref{figure:cat_dist_app}.

\section{Negative Sampling Strategies}\label{appendix:negative}
In our preliminary experiments, we tried several negative sampling strategies before arriving at the one described in \S~\ref{step-inference-task} of the paper. \\
\textbf{Random Strategy}: Each negative example is a randomly sampled step or goal from another article. With high probability, it is indeed incorrect with regard to the current prompt: the sampled step cannot be used to accomplish the current goal, or the sampled goal cannot be accomplished by the current steps. Empirically, data sampled with this approach makes the tasks too easy for both models and human accuracy, as the negative examples are too irrelevant. As an example for Step Inference, given the goal ``Play Guitar'' and the positive step ``practice basic scales'', the negative candidates are ``buy a used car'', ``gather the ingredient``, and ``wind down with meditation''.\\
\textbf{Other Keyword-KNN Strategies}: Instead of taking the embedding of the verbs in each goal or step phrase, we consider nouns, concrete part-of-speeches and all words. Empirically, data sampled with this approach includes many ambiguous questions, as many negative candidates have identical meaning to the positive candidate. As an example considering all words for Step Inference, given the goal ``Play Guitar'' and the positive step ``practice basic scales'', the negative candidates are ``play basic scales'', ``learn the scales``, and ``learn basic chords''.\\
\textbf{Masked Language Model Strategy}: We also experiment with using Masked Language Modeling (MLM), BERT's pre-training task, to generate (instead of sampling) negative candidates from the positive one. Given the positive candidate, we iteratively mask out a random token and ask BERT to predict the most likely token different than the original one. For example, after several such iterations, a step like ``read your local phone book" could become ``find your local history book", ``read your favorite story book", or ``call my own phone number'', which would be the three negative candidates. The idea is to use BERT as an adversary for subsequent models, by generating negative candidates that have high MLM likelihood and therefore make the examples challenging. Empirically, however, it turns out that such an adversary is imperfect and can be easily conquered by models; moreover, the iterative prediction process is too time-consuming to scale.

\begin{figure*}[t!]
    \centering
    \includegraphics[width=\linewidth]{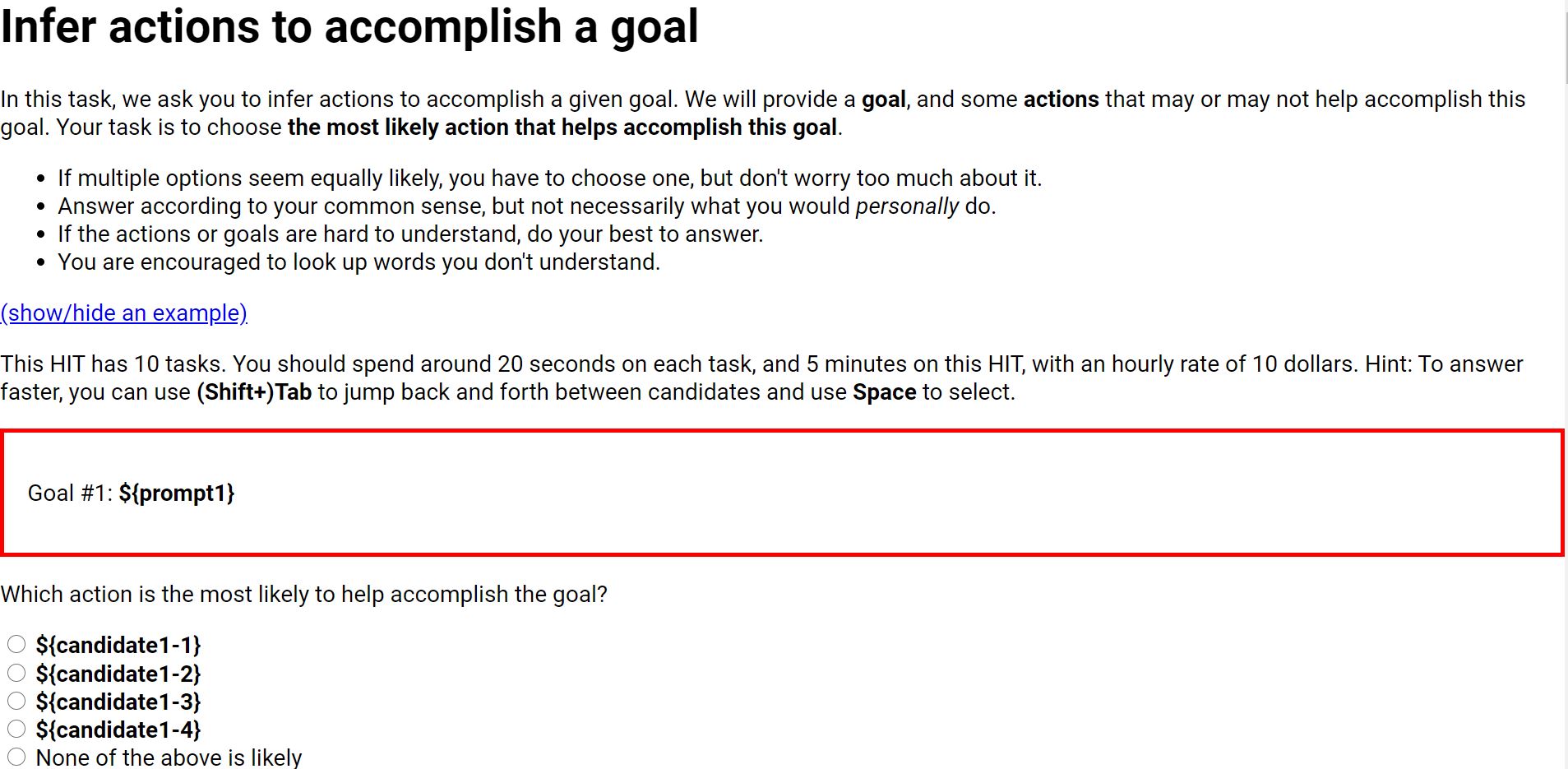}
    \caption{Screenshot of the HIT design for the Step Inference task.}
    \label{step-hit}
\end{figure*}

\section{Quality Control Filters}\label{appendix:filters}
As described in \S~\ref{step-inference-task} and \S~\ref{goal-inference-task}, we apply a collection of hand-crafted filters to the automatically generated examples to remove low-quality ones. The details of each filter are as follows:\\
\textbf{Category filter}: We remove examples involving articles under certain wikiHow categories. The categories we leave out are either too obscure (e.g. Astrology Relationships) or require expert domain knowledge to reason about (e.g. Car Engine Repairs), with the hope that the remaining categories contain more what we would call ``common sense" knowledge that an average human has.\\
\textbf{Lexical-Overlap filter}: We remove examples where there is a lexical overlap between the prompt and each candidate. We exclude stopwords and lemmatize each word using spaCy before computing the overlap. \\
\textbf{TF-IDF filter}: We remove examples with overly uninformative prompts or candidates. We exploit TF-IDF as a proxy for how indicative a certain step is of the article it comes from. The motivation is that in Step Inference, for example, given a prompt step, the task is to choose its corresponding goal; then a prompt step like ``gather your materials'' is almost not informative at all for humans/models to tell which goal it serves, as a large number of articles may include a step like this. Thus, we treat each wikiHow article as a document and calculate the TF-IDF of each token, and only retain steps that have at least one token whose TF-IDF value surpasses a certain threshold.\\
\textbf{Length filter}: We remove examples with overly short prompts or candidates. The motivation is similar to that of the TF-IDF filter, i.e. too short goals/steps may not be informative enough to make a clean example. For example, steps like ``Finished!'' or ``Serve!'' are hard to tell apart if one of them is the positive candidate while the other is negative. To reduce such kind of noise in the automatically generated examples, we filter out steps/goals that are shorter than a specific threshold.\\
\textbf{Similarity filter}: We use similarity-based filters to remove examples where some negative candidate is also likely to be a plausible answer. The similarity scores are calculated using cosine similarity between BERT embeddings described in \S~\ref{step-inference-task}. In Step Inference, we set an upper threshold on the similarity between any negative step and any step from the prompt goal, with the motivation that negative steps should not serve the prompt goal. For Goal Inference, likewise, we ensure that the similarity between the prompt step and all steps from any negative goal is lower than a threshold, thus trying to minimize the cases where the prompt step also helps achieve negative goals.

\section{Crowdsourcing Details}\label{appendix:crowdsourcing-details}

Some noise exists in our automatically generated examples, because some of them do not have a single correct answer. This can happen when a sampled negative candidate is in fact correct. For example, in the Goal Inference task, consider an example where the give step is ``practice swings'', the expected positive candidate step is ``Play Golf'', and a candidate negative example is ``Play Drums''. ``Play Drums'' is sampled due to its high embedding similarity with ``Play Golf'' and is also a reasonable goal for ``practice swings (of the drumsticks)''. This is an ambiguous example and should be excluded from the test set, which is supposed to be a benchmark for models. Hence, we ask crowd workers to validate a subset of the examples. An example is shown in Figure~\ref{step-hit}.

We perform crowdsourcing on Amazon Mechanical Turk, requiring Master Qualification and a lifetime HIT approval rate over 90\% for the crowd workers. 

For each of Step Inference and Goal Inference, we randomly sample 4,800 examples as input, and for each example we ask 3 crowd workers from Amazon Mechanical Turk to choose the most likely candidate. Every HIT includes 15 examples with a pay of \$0.83, estimated to be completed in 5 minutes, equivalent to an hourly rate of \$9.96.

For Step Ordering, we randomly sample 9,300 examples, and for each example we ask 3 crowd workers to order the events. Every HIT includes 30 examples with a pay of \$0.83, estimated to be completed in 5 minutes, equivalent to an hourly rate of \$9.96. 

In the test set, we retain only examples where all 3 crowd workers agree on the correct answer. See Table~\ref{crowd_yield} for the distribution of annotators' agreement with the gold labels and the final yield rate (i.e. proportion of examples with all 3 workers answering correctly). 
\begin{table}[t!]
\small
\centering
\begin{tabular}[t]{cllll}
\toprule
 \makecell{\# Wokers selecting\\ expected positive} & \makecell{Step\\Infer.} & \makecell{Goal\\Infer.} & \makecell{Step\\Ordering} \\ \midrule
0 & 307 & 534 & 1,014 \\
1 & 556 & 813 & 1,854 \\
2 & 1,031 & 1,122 & 2,732 \\
3 & 2,250 & 1,703 & 3,100 \\ \midrule
Yield rate & .543 & .408 & .356 \\ 
\bottomrule
\end{tabular}
\caption{The distribution of agreement with the gold labels and the yield rate for each task. }
\label{crowd_yield}
\end{table}

\section{Harder and Noisier Benchmarks}\label{appendix:subbenchmarks}

As described in \S~\ref{crowdsourcing-validation}, only examples where all 3 crowd workers choose the correct label are kept in the benchmarks to ensure high quality. We also release the \textit{sub-benchmarks} including examples where 2 out of 3 workers choose the correct label. Naturally, these sets include both examples that require more attention to answer correctly and those that are inherently ambiguous, which we cannot distinguish at present. The performance of some state-of-the-art transformer models are shown in Table~\ref{sub-performance}. 

\begin{table}[t!]
\small
\centering
\begin{tabular}[t]{lllll}
\toprule
 & \makecell{Step\\Infer.} & \makecell{Goal\\Infer.} & \makecell{Step\\Ordering} \\ \midrule
Train size & 404,057 & 239,239 & 841,317 \\
Test size & 1,031 & 1,122 & 2,732 \\ \midrule
BERT & .731 & .563 & .681 \\
XLNet & .750 & .623 & .680 \\ 
RoBERTa & .789 & .623 & .692 \\ \midrule
Crowd workers & .67 & .67 & .67 \\
\bottomrule
\end{tabular}
\caption{The accuracy of state-of-the-art models on the sub-benchmarks, finetuned on the training sets. }
\label{sub-performance}
\end{table}

\section{More Open-Ended Examples}\label{appendix:qualitative}
In addition to the examples in \S~\ref{open-ended}, we provide more open-ended examples for each task here. 

\subsection{Step Inference}

For these open ended examples, our Step Inference model is trained in a 100-choose-1 format with 99 negative samples, instead of 4-choose-1, given 3 steps instead of 1. During evaluation, we use the softmax value in the final layer as the probability for each candidate. We rank the probabilities and report the top 3. Here are some more examples: 

\noindent
{\bf Input goal:}  Choose a Role Model \newline
{\bf Predicted steps:} learn about their successes and failures (correct), show interest in their lives, ask about their life \\

\noindent
{\bf Input goal:}  End a Letter of Apology\newline
{\bf Predicted steps:} use a signature that conveys your emotions (correct), try to personalize the letter as much as possible, focus on the facts of the situation

\subsection{Goal Inference}

For Goal Inference, we follow the same procedure as above. Here are some more examples: 

\noindent
{\bf Input steps:}  buy or rent a good hammer drill, drill a pilot hole, insert a high quality masonry drill bit \newline
{\bf Predicted goals:} Drill Into Concrete (correct), Drill Holes Through Glass, Dig a Hole \\

\noindent
{\bf Input steps:}  cultivate a memorable persona, keep an equal balance between your vlogging and your work life. review your channel\newline
{\bf Predicted goals:} Become a YouTube Guru (correct), Become a Film Buff, Become a Videographer

\subsection{Step Ordering}

For Step Ordering, the model can perfectly order the steps in many wikiHow articles unseen during training. To perfectly order an article, the model needs to correctly order all possible pairs of steps in an article. Here are 2 example articles with 10 steps: 

\noindent
\textbf{Change Your Name of a Minor in Colorado}: (1) make sure the child is eligible for a name change,  (2) choose the right court,  (3) download and review your forms, (4)  get a fingerprint-based criminal background check,  (5) complete the necessary forms,  (6) get consent from the non-custodial parent,  (7) file your petition with the appropriate court,  (8) serve the non-custodial parent, (9) publish the proposed name change,  (10) attend the hearing on your petition.\\ \ \\
\textbf{Draw a Simple Teddy Bear}:  (1) draw a circle for the teddy bear’s head and an oblong for its body,  (2)  add two curved lines on each side of the oblong for the bear’s arms,  (3)  draw two small circles below the oblong for the bear’s feet, (4)  add the ears using two small circles on each side of the head,  (5) draw details of the face,  (6) add details on the bear’s pads using three small circles and a bean shape below it,  (7) draw a shirt for the bear,  (8) make the bear look furry by using small strokes in drawing its body,  (9) erase unnecessary lines,  (10) color the drawing.

\end{document}